\documentclass{article}

     \usepackage[preprint]{neurips_2021}

\usepackage[utf8]{inputenc} % allow utf-8 input
\usepackage[T1]{fontenc}    % use 8-bit T1 fonts
\usepackage{hyperref}       % hyperlinks
\usepackage{url}            % simple URL typesetting
\usepackage{booktabs}       % professional-quality tables
\usepackage{amsfonts}       % blackboard math symbols
\usepackage{nicefrac}       % compact symbols for 1/2, etc.
\usepackage{microtype}      % microtypography
\usepackage{xcolor}         % colors
\usepackage{natbib}   
\usepackage{graphicx}
\usepackage{subfig}
\usepackage{amsmath}
\usepackage{blindtext}
\usepackage[ruled,vlined]{algorithm2e}

\usepackage{multirow}

\SetKwInput{KwInput}{Input}
\SetKwInput{KwOutput}{Output}

\newcommand{\be}{\begin{equation}}
\newcommand{\ee}{\end{equation}}

\title{Tourbillon: a Physically Plausible Neural Architecture} 
\author{
 Mohammadamin Tavakoli \\
  Department of Computer Science\\
  University of California, Irvine\\
  \texttt{mohamadt@uci.edu} \\
  
  \And
  Peter Sadowski\\
  Department of Information and Computer Sciences\\
  University of Hawai‘i at M\={a}noa\\
  \texttt{peter.sadowski@hawaii.edu} \\
 
  \And
  Pierre Baldi\\
  Department of Computer Science\\
  University of California, Irvine\\
  \texttt{pfbaldi@uci.edu} \\

}

\begin{document}

\maketitle

\begin{abstract}
In a physical neural system, backpropagation is faced with a number of obstacles including: the need for labeled data, the violation of the locality learning principle, the need for symmetric connections, and the lack of modularity. Tourbillon is a new architecture that addresses all these limitations. At its core, it consists of a stack of circular autoencoders followed by an output layer. The circular autoencoders are trained in self-supervised mode by recirculation algorithms and the top layer in supervised mode by stochastic gradient descent, with the option of propagating error information through the entire stack using non-symmetric connections. While the Tourbillon architecture is meant primarily to address physical constraints, and not to improve current engineering applications of deep learning, we demonstrate its viability on standard benchmark datasets including MNIST, Fashion MNIST, and CIFAR10. We show that Tourbillon can achieve comparable performance to models trained with backpropagation and outperform models that are trained with other physically plausible algorithms, such as feedback alignment.
\end{abstract}

\section{Introduction}

In physical neural systems, whether biological or neuromorphic, backpropagation is faced with a number of problems (e.g. \cite{bengio2015towards, nokland2016direct,baldi2016local,guerguiev2017towards,baldi2018learning, lillicrap2020backpropagation}.
Of greatest relevance here are the following problems: 
\begin{enumerate}
\item Labeling: the need for large quantities of labeled data to compute gradients for supervised learning.  
\item Locality: in a physical neural system, the learning rule for adjusting the synaptic weights must be local, i.e. depend only on variables that are available locally, both in space and time, at the synapse. 
\item
Symmetry of Connections: backpropagation requires  
precisely symmetric connections between the forward and backward passes that may be hard to realize in a physical neural system.

\item Distances: backpropagation requires propagating signals over significant neural distances, which could lead to signal dilution, and distorted or unstable gradients. 
\item Developmental Modularity:
 backpropagation in general requires having a complete architecture in place before training can begin, which may not be suitable under certain developmental constraints.
 
\end{enumerate}

A number of solutions have been suggested to try to address these problems, in isolation or in small combinations, but no approach addresses all of them at once well. The Tourbillon architecture  proposes to address all of them at once by combining different ideas including stacked autoencoders, random backpropagation, and recirculation.
We stress that the goal here is to address the problems above in  {\it physical} neural systems, and not to derive a new architecture or algorithm that is practically useful for current engineering applications of deep learning.

\section{Stacked Autoencoders}

One well known approach for dealing with the Labeling problem is to use a stack of autoencoders \citep{hinton2006fast, baldi2012autoencoders}, where each autoencoder in the stack is trained in self-supervised manner to reproduce the hidden representation produced by the previous autoencoder in the stack. The intuition is that such a stack can learn increasingly more abstract and powerful representations of the input data without the need for labels. Labels can be used at the top of the architecture to train the top layer in a supervised manner by stochastic gradient descent, with the additional option of backpropagating through all the layers to fine-tune the entire stack (although there is some debate in the field as whether the latter is helpful or not \citep{tschannen2018recent}).
\iffalse
The following equation shows a stack of two autoencoders where $H_i$ shows the hidden representations and $E_i$ and $D_i$ represents the encoder and decoder neural networks.
\be
    H_1 = E_1(x_1), \hat{x}_1 = D_1(H_1),\quad
    H_2 = E_2(H_1), \hat{H}_1 = D_2(H_2)
\label{eqn:tourb}
\ee
\fi
The stacked-autoencoder approach has the potential for addressing two other problem: Distance and Developmental Modularity.
Because backpropagation occurs inside each autoencoder in the stack, error gradients are potentially propagated over shorted distances confined to the size of the individual autoencoders. Furthermore, individual autoencoders can be trained at least as soon as all the previous autoencoders in the stack are trained and possibly earlier, avoiding the need to wait for the entire architecture to be wired before training can begin.

The stack of autoencoders approach however does not address the problems of Locality and Symmetry of the forward weights. Each autoencoder is deep by definition, in the sense that it has at least one hidden layer, and therefore it requires applying the backpropagation algorithm across at least two adaptive layers of weights. This in turn requires having a deep learning channel (i.e. "wires") for transmitting the error signal and making it local to the hidden layer(s). Furthermore, the weights on these wires have to be symmetric to the forward weights in order to implement standard backpropagation.

\iffalse
\cite{baldi2016local}
 \cite{baldi2018learning}
 recirculation is RBP
 \cite{baldiRBP2016AI}
 RBP and DLC
 \cite{baldi2017learning}
 symmetries of the DLC
 \cite{hinton1988learning}
 \cite{o1996biologically} recirculation
 
 \cite{lillicrap2016random,baldiRBP2016AI,baldi2017learning}
\fi
\section{Random Backpropagation}

Feedback alignment (FA) or random backpropagation (RBP) refers to a family of algorithms
\citep{lillicrap2016random,baldiRBP2016AI} that address the Symmetry problem by using non-symmetric random weights in the backward pass. For a forward weight $w_{ij}$, the backpropagation learning equation can be written as:
\be
\Delta w_{ij}=\eta B_i^{post} O_j^{pre}
\label{eq:BP}
\ee

where $\eta$ denotes the learning rate, $B_i^{post}$ denotes the postsynaptic backpropagated error, and $O_j^{pre}$ denotes the presynaptic activity. In contrast, RBP can be expressed as:

\be
\Delta w_{ij}=\eta R_i^{post} O_j^{pre}
\label{eq:RBP}
\ee

where $R_i^{post}$ denotes the postsynaptic, randomly backpropagated, error.
Many flavors of RBP can be obtained by using different notions of randomness to select the backward weights, by including various skipped backward connections, by making the backward weights adaptive, and so forth \citep{baldiRBP2016AI,baldiSymmetries2017}.
\iffalse
For instance, direct feedback alignment (DFA) \citep{nokland2016direct,}, as a crude approximation of the feedback alignment, backpropagates the error signal obtained at the top layer to each of the lower layers independently using fixed random backward weights as follows:
For layer $i$ in a neural network with $k$ layers, the activation will be $O^i$, where $f$ is the nonlinear activation function. Obtaining the error $T-O^{k}$ at layer $k$, DFA derives the post-synaptic error term $\delta O^i$ for $W^i$ using the random matrix $M^i$.
\be
O^i = f^i(W^iO^{i-1}), \quad \delta O^i = M^i(T-O^k)\odot f^{\prime}(O^i)
\label{eqn:dfa}
\ee

And updates the weight $W^i$ using the pre-synaptic activation:
\be
\delta W^i = -\delta O^i(O^{i-1})^T = M^i(T-O^k)\odot f^{\prime}(O^i)(O^{i-1})^T
\ee
\fi
While RBP addresses the Symmetry problem, by itself it does not address the other problems.
Furthermore, while RBP simulations have succeeded on the MNIST and CIFAR benchmark data sets, it has been noted that RBP algorithms do not work well with convolutional layers \citep{bartunov2018assessing, moskovitz2018feedback, refinetti2020dynamics}. 
\iffalse
It has been noted that RBP algorithms have difficulties with convolutional layers \citep{liao_backpropagation_2015, moskovitz2018feedback, refinetti2020dynamics}.
\fi
A few methods have been proposed to address this apparent weakness of RBP algorithms, however, most of them are not biologically plausible. For instance, \citet{liao_backpropagation_2015} proposes to use uniform sign-concordant feedback (uSF), where each forward matrix $W$ in the forward pass is replaced by the matrix $sign(W)$ in the backward pass.
They also pointed that normalization methods such as batch normalization and batch Manhattan can improve the performance of RBP when dealing with convolution layers. \citet{moskovitz2018feedback} modifies the uSF method by putting additional constraints on the backward weight matrices. In all these approaches, the backward weight matrices must have some knowledge about the forward matrices.

\iffalse
, all of which require access to the forward weight matrices. We suspect that one of the main reasons for this is that each neuron in the same convolutional layer receives a different randomly backpropagated error. When the weights are updated, all these randomly backpropagated errors must be averaged to produce the final update that is applied simultaneously to all the corresponding weights. This averaging of random backpropagated errors may produce a signal that is close to 0, at least on average, and thus not very effective for learning. Utilizing a similar idea, \citet{bartunov2018assessing} breaks the weight sharing constraint in the convolutional networks and develop locally connected networks to scale up the biologically plausible architectures. Although this is a promising approach, explaining it is out of the scope of the presented manuscript. In addition to the procedures that have been proposed in the literature to overcome this problem here we propose two additional approaches. 
\fi

\section{Circular Autoencoders and Recirculation Algorithms}
\label{learninig}
\begin{figure}[ht]
\begin{center}
\includegraphics[width=1.0\columnwidth]{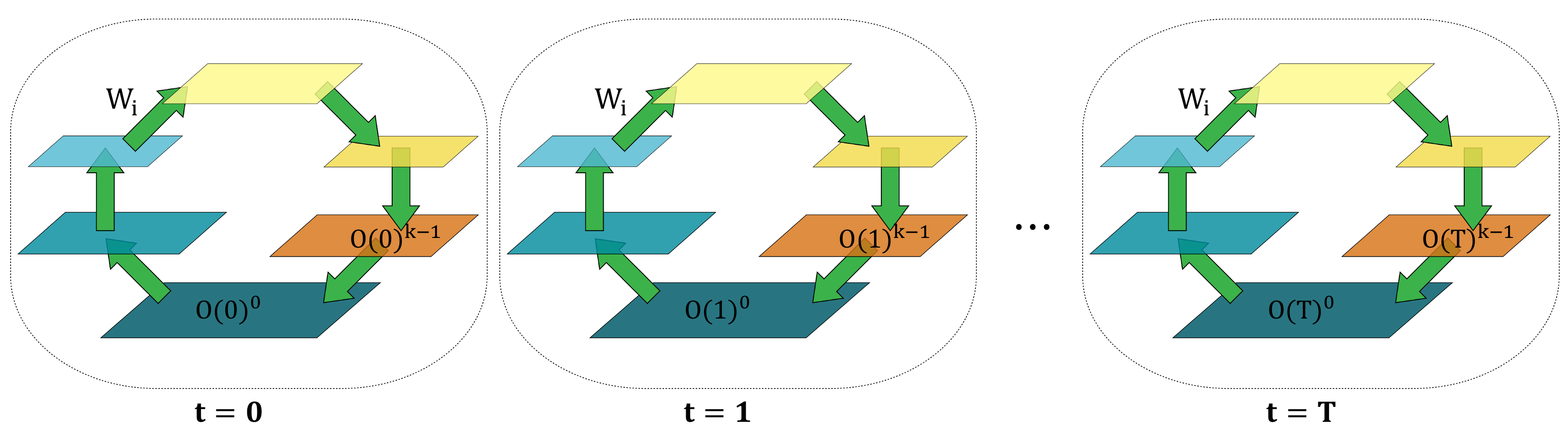}
\end{center}
\caption{Circular autoencoder. The output layer (layer $k$) is equal to the input layer (layer 0). For a given input, information can be propagated (recirculated) multiple times through the autoencoder loop. The first pass is indexed by 0, the last one by $T$. For instance, $O(0)^{k-1}$ denotes the output of layer $k-1$ during the first circulation.}

\label{fig:circular}
\end{figure}

In a standard feed forward autoencoder, the data itself provides the targets (self-supervised learning). The data and hence the targets are available in the input layer. However, they are not available in the output layer, in the sense that they are not physically local to the output layer. This problem is addressed by circular autoencoders where the output layer is physically equal (or physically adjacent) to the input layer (Figure \ref{fig:circular}).
With the circular layout, targets and errors can be computed at the level of the input/output layer.
Standard backpropagation, or even standard RBP, of these targets would require a channel (wires) running backward from the output layer to the hidden layer. However, because of the circular layout, one may suspect that it may be possible to use the forward connections to propagate target and error information during learning. This is the fundamental idea behind recirculation, a family of algorithms for training circular autoencoders that do not require backward connections \citep{hinton1988learning,o1996biologically, baldi2018learning}.

Consider a circular autoencoder with layers numbered from $0$ to $k$, where $0$ and $k$ correspond to the input layer and use the index $t$ to denote different feed forward passes through the autoencoder, with the first pass indexed by $t=0$. After the first pass, one can {\it locally} compute the error $T-O^k(0)$, where $T$ is the target, i.e. the original data in the case of an autoencoder.
This error could be used to train the top layer of the circular autoencoder by gradient descent, and then train the other layers by using a form of RBP where the error signal is obtained by propagating the error $T-O^k(0)$ using the forward weights of the circular autoencoder. This however requires propagating two different kinds of signals, activities, and errors, through the circular autoencoder. Thus rather than recirculating the error, a more uniform approach can be obtained by recirculating activities. If $O^i(t)$ denotes the output of layer $i$ during the forward pass indexed by $t$, the main idea behind the recirculation family of algorithms is to use 
$O^i(t)$ as the target for the output $O^i(t')$ taken at a later time $t'$. The intuition behind this is that the data may become increasingly corrupted as it is being recycled.
Several different algorithms can be obtained, by varying, for instance, the choices of $t$ and $t'$ and the manner in which outputs from different passes are combined (e.g. convex combinations) \citep{baldi2018learning}. Here we will use the simplest form, which often works the best, based only on the plain activities of the first two passes corresponding to $t=0$ and $t=1$. 
Thus, for any weight $w_{ij}$ in the circular autoencoder, the corresponding learning-by-recirculation equation becomes:

\be
\Delta w_{ij}=\eta [O_i(0)-O_i(1)]^{post} [O_j(0)]^{pre}
\label{eqn:rec}
\ee

This rule has a Hebbian-product form between a post-synaptic term and a pre-synaptic term, and is similar to backpropagation, except that the postsynaptic backpropagated error is replaced by the postsynaptic recirculation error $[O_i(0)-O_i(1)]^{post}$.
This error term is local in space and is assumed to be also local in time. This requires the assumption that two consecutive passes through the circular autoencoder fall within the time window that defines locality in a particular physical system. For the input layer, the vector $O^0(0)$ corresponds to the input data, hence also to the targets in the case of an autoencoder. Thus the recirculation learning equation for the top layer of weights is exactly equal to backpropagation (stochastic gradient descent ). Having stochastic gradient descent applied to the top layer is also essential for RBP to work. The recirculation error $[O_i(0)-O_i(1)]$ can also be thought of as a derivative of the activity. Thus, in short, recirculation learning rules rely on the product of the pre-synaptic activity by some measure of change in the post-synaptic activity, which is used to communicate error information. 
Although in this work we are not using spiking neurons, such learning rules are closely related to the concept of spike time-dependent synaptic plasticity (STDP). STDP Hebbian or anti-Hebbian learning rules have been proposed using the temporal derivative of the activity of the post-synaptic neuron \citep{NIPS1999_1658} to encode error derivatives. 
 \iffalse
 \cite{baldi2016local}
 \cite{baldi2018learning}
 recirculation is RBP
 \cite{baldiRBP2016AI}
 RBP and DLC
 \cite{baldi2017learning}
 symmetries of the DLC
 \cite{hinton1988learning}
 \cite{o1996biologically} recirculation
 
 \cite{lillicrap2016random}
 \fi
 In the Appendix, some of the simulations also explore a slightly different anti-Hebbian learning rule:

\be
\Delta w_{ij} =-\eta
[O_i(0)-O_i(1)]^{post}[O_j(0)-O_j(1)]^{pre}
\label{eqn:new_lr}
\ee

where the pre- and post-synaptic terms are more similar in form. This rule is applied to all the layers, but the top adaptive layer where the standard SGD rule is used.

\section{Tourbillon}
We propose the Tourbillon architecture as a stack of circular autoencoders followed by a partially or fully-connected layer between the hidden representation of the top circular autoencoder and the output layer. Each circular autoencoder is broken down into two components, the encoder, and the decoder components. The hidden layer that is shared by the encoding and decoding components of a  circular autoencoder is called the hinge hidden layer. In the stack, the hinge hidden layer for the circular autoencoder at level $i$, becomes the input layer for the next circular autoencoder at level $i+1$ in the stack.

The circular autoencoders are trained by recirculation (self supervised manner) and the top layer can be trained in a supervised manner by stochastic gradient descent, with the option of continuing the error propagation through the entire stack, as a form of RBP using non-symmetric connections.
In a physical neural system, this requires that the targets be locally available at the final output layer.
This architecture addresses all five problems mentioned in the introduction. Most of the training is unsupervised, all the learning rules are local in space and time, the forward and backward connections are not symmetric, information is propagated only over relatively short distances, the development is modular, and also convolutions can be incorporated into this framework. To prove its viability, we test this architecture on several benchmark problems in the Experiments section.

\begin{figure}[ht]
\begin{center}
\includegraphics[width=0.8\columnwidth]{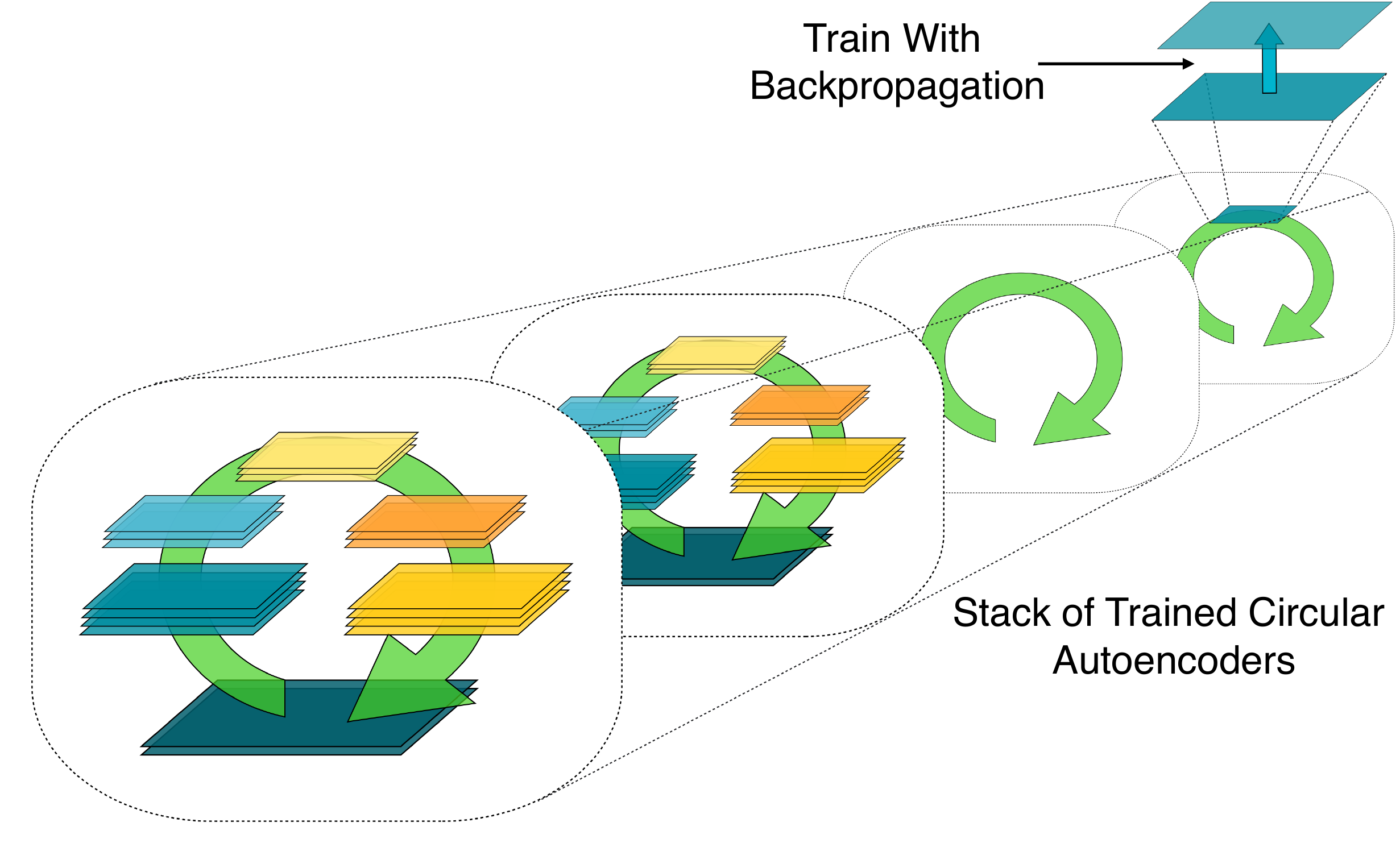}
\end{center}
\caption{Tourbillon architecture with a stack of circular autoencoders trained by recirculation. Some layers are drawn with multiple sheets to signify the possible presence of convolutions.
}
\label{fig:tourb_sheet}
\end{figure}

Although we simulate Tourbillon in its most basic version, there are a number of possible variations on the basic Tourbillon idea and training algorithms. In particular, in the basic version, the stack of autoencoders is trained sequentially bottom-up starting from the initial input layer. It is possible to consider more asynchronous modes, where for instance any circular autoencoder in the stack is being trained as soon as it receives an input from a lower autoencoder.

\subsection{Fine-Tuning Using the Decoders' Connections to Create a Deep Learning Channel}
\label{finetune} 

Once the stack of circular autoencoders is in place, possibly after a developmental phase, the connections of their decoder components provide a deep learning channel, i.e. a physical channel through which error obtained at the top of the architecture could be communicated to the lower layers. This channel, whose weights are initially learned by recirculations, can be used to fine-tune the entire Tourbillon architecture, using a form of random backpropagation, since there is no imposed symmetry between the weights of the encoder and decoder networks of each circular autoencoder. This produces a kind of RBP error signal for the hinge hidden layer of each circular autoencoder. If the encoder component of circular autoencoders has multiple layers, different possibilities can be considered for the fine-tuning phase when the error signal arrives at the hinge. In all the simulations, we use circular autoencoders with a single hidden layer, and thus the RBP error that arrives at the hinge is local to the corresponding encoder weights and these can be fine-tuned by standard RBP (Equation \ref{eq:RBP}).
 
 \iffalse
 This process is well defined if the circular autoencoders have a single hidden layer, which is what we use in all the simulations. When there are multiple layers in the encoder and decoder components of each circular autoencoder, then a number of different possibilities can be considered.
 
 However these seem to require the presence of additional connections in order to maintain locality. For instance if the encoder and decoder networks have symmetric architectures (e.g. same number of layers, same layer sizes) then the RBP error in each layer of the decoder could be used to train the corresponding layer on the encoder side. This alone would require additional ``horizontal'' connections running  from the decoder to the encoder to propagate the error signal.
\fi
\subsection{Compositionality of the Tourbillon Architecture}
\label{toubillonization}
Due to the modularity of the Tourbillon architecture, it can be composed in many different ways. In particular, it can be utilized to build a physically plausible twin architecture for every physically non-plausible neural network trained with backpropagation or random backpropagation.
%%with non-local connections. 
Although the physically plausible architecture does not outperform its twin, this idea opens the door for further studies of physically plausible approaches. Here we explain the process of building the twin architecture through two examples: (1) Tourbillon U-Net; and (2) Tourbillon recursion. We also provide the general algorithm of building the physically plausible twin architecture in Algorithm \ref{alg:tourb}. 

The first example is the ``tourbillonization'' of a U-Net \citep{ronneberger2015u} architecture. The U-Net architecture is basically a feed forward autoencoder with a multi-layer encoder, and a multi-layer decoder, as well as additional skip connections (which we ignore for simplicity). This architecture was originally developed for image segmentation problems, where the segmented images are used as the targets. Each layer of a U-Net architecture can be replaced by a circular autoencoder to build a Tourbillon version of U-Net. The only non-local aspect of the Tourbillon U-Net is the need for targets in the final output layer. This could be addressed by turning U-Net itself into a circular autoencoder, which leads to the second example where the Tourbillon approach can be used recursively, by replacing each hidden layer of a circular autoencoder, with a circular autoencoder, a recursion that in principle could be iterated several times. In the later sections we experiment the tourbillonizaiton of U-Net, however, how a Tourbillon of Tourbillons architectures can be trained efficiently is left out for future work. 
\iffalse
\textcolor{red}{Probably the sentence above should be moved to the Discussion. Amin: should you try to draw a tourbillon of tourbillons?}
\fi

\begin{algorithm*}[h]
\SetAlgoLined
\SetKwFunction{FMain}{train\_circular\_ae}
\SetKwFunction{Frec}{recirculation}
\SetKwProg{Fn}{Function}{:}{}
\KwInput{A network $M$ with $m$ sequential layers $L=[L_1, ..., L_m]$, $data$: training data, and \Frec($c$) function which trains a circular autoencoder $c$.}
\KwOutput{The physically plausible twin $M^\prime$.}
 initialization; $L^\prime$: an empty list\\
 \For{$i \gets 1$  \KwTo $m$}{
    \eIf{$L_i$ has learnable parameters}{%
            $l$ = \FMain{$[L1, ..., L_i]$}\\
            $L^\prime$.append($l$)
        }
        {
            $L^\prime$.append($L_i$)
        }
    }
  $M^\prime$ = $[data]$\\
  \For{$i \gets 1$  \KwTo $m$}{
    $M^\prime$ = $L^\prime_i$($M^\prime$)
  }
  \Fn{\FMain{$[L_1, ..., L_i]$}}{
        input = $L_{i-1}(...(L_1(data)))$\\
        circular\_ae = $L_i^{-1}(L_i(\text{input}))$\\
        \Frec(circular\_ae)\\
        \KwRet $L_i$\\
  }
 \caption{Tourbillonization}
 \label{alg:tourb}
\end{algorithm*}

\section{Experiments}
\label{exp}
In the experiments, we train circular autoencoders and Tourbillon architectures capable of supporting feed forward fully-connected layers, 2D convolutions, 2D max pooling, 2D up sampling, reshaping, and five different non-linear activation functions. %The code will be made available through GitHub upon acceptance. 

\subsection{Data}
In the experiments, we use three well-known benchmark datasets: (1) \textbf{MNIST}, comprising 70,000 gray-scale images of handwritten digits of size $28 \times 28$ with each pixel normalized to the $[0, 1]$ range. We used the normal 60000 and 10000 train-test split; (2) \textbf{Fashion MNIST} comprising 70,000 gray-scale images of fashion accessories of size $28 \times 28$ with each pixel normalized to the $[0, 1]$ range. We used the normal 60000 and 10000 train-test split; and (3) \textbf{CIFAR-10} comprising 60,000 images of size $32 \times 32$ with three RGB channels corresponding to 10 object classes with each pixel normalized to the $[0, 1]$ range. We used the normal 50000 and 10000 train-test split. For the following experiments, the models are trained using a single NVidia Titan X GPU.

\subsection{Tourbillon Building Blocks}
\label{circular_ae}
\begin{figure}[h]
\begin{center}
\includegraphics[width=1.0\columnwidth]{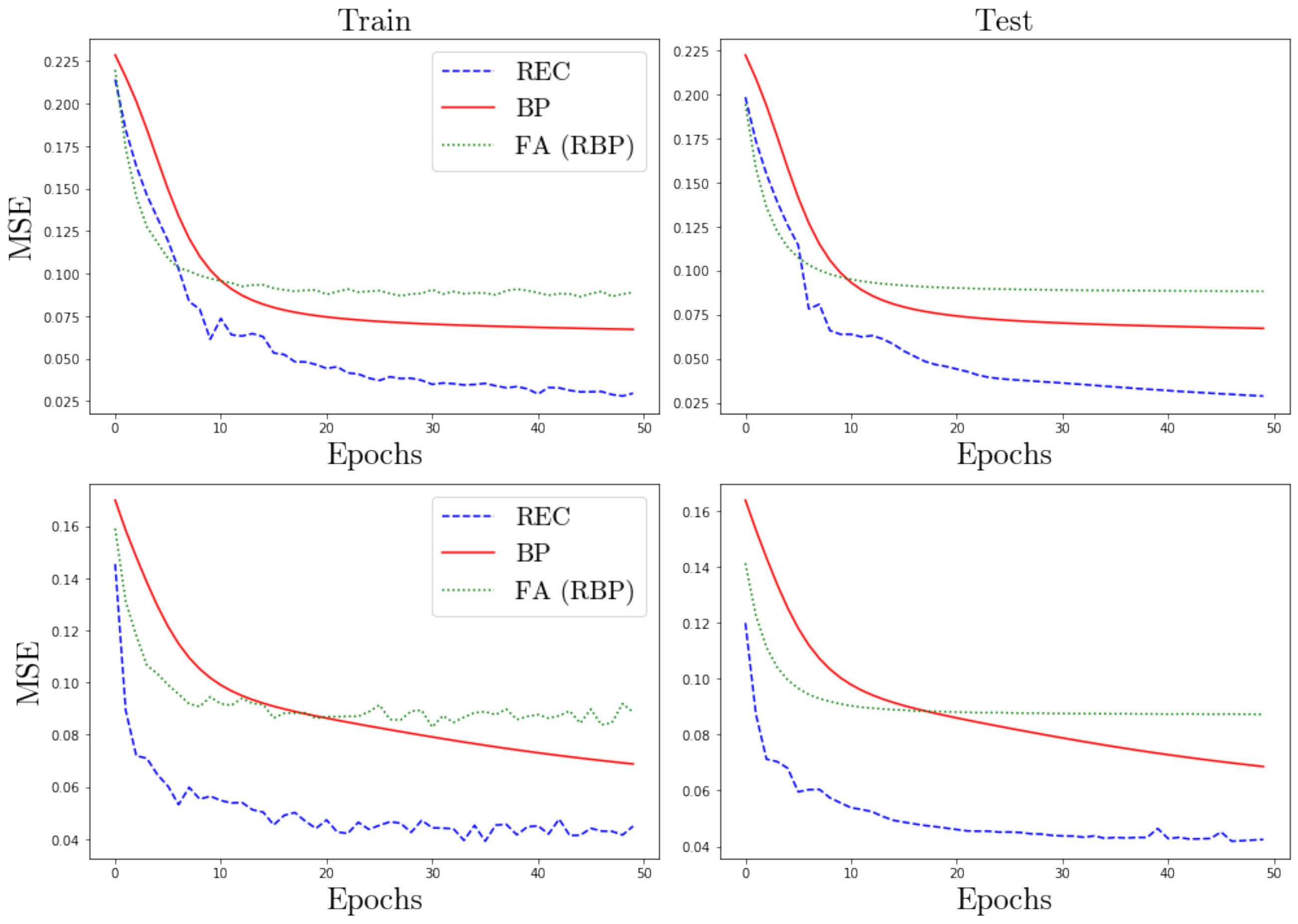}
\end{center}
\caption{Comparison of the training and test mean-squared reconstruction loss of three equivalent autoencoders trained using the MNIST and Fashion MNIST datasets. The first row shows the performance of the models for the MNIST dataset, and the second row shows the performance for the Fashion MNIST dataset. REC corresponds to a circular autoencoder trained with recirculation, FA corresponds to the same architecture trained with feedback alignment, and BP corresponds to the same architecture trained with backpropagation.}
\label{fig:mnist_fashion_mnist_recirc}
\end{figure}

Since the building blocks of the Tourbillon architecture are circular autoencoders, we train compressive circular autoencoders using the recirculation algorithm (Equation \ref{eqn:rec}) and compare the mean-squared reconstruction loss (similar results are obtained with the relative entropy loss) with that of the corresponding architecture trained with both backpropagation and random backpropagation. For experiments with MNIST and Fashion MNIST, the circular autoencoder consists of two feed forward fully-connected layers, where the middle layer is of size 256.
For experiments with CIFAR10, the circular autoencoder consists of two 2D convolution layers with kernels of size $(5 \times 5)$ to construct a hidden representation of size $(32 \times 32 \times 6)$ in the hinge layer. Both convolutional layers use strides of size $(1 \times 1)$ with zero padding to preserve the size of the input. 
Figure \ref{fig:mnist_fashion_mnist_recirc} shows the training and test error curves for the MNIST and Fashion MNIST datasets. 
Figure \ref{fig:cifar}, first row, shows the training and test error curves for the convolutional circular autoencoder trained using the CIFAR10 dataset. In all these experiments, recirculation leads to reconstruction errors that are comparable, and possibly even better, than the corresponding errors obtained using backpropagation and random backpropagation.
In all these experiments, the models are trained with a mini-batch of size 64 to minimize the reconstruction error. Across all three models, all hidden layers use a \textit{tanh} activation function and the final layer uses a \textit{logistic} activation function (since the pixels are normalized to $[0,1]$). During our experiments, we observed that the presence of non-linear activation functions in intermediate layers is essential for learning, especially for the circular autoencoder which is consistent with the results provided in \citet{baldi2016learning}. For all models, a grid search was used to optimize the hyperparameters. As a result, models trained with backpropagation and random backpropagation across all three datasets use a starting learning rate of 0.01 for all the layers, with a momentum term of 0.8. On the other hand, for the circular autoencoder traioned using MNIST and Fashion MNISt datasets, the grid search resulted in starting learning rates of 0.01 and 0.001 for the first and second layers respectively. For the circular autoencoder trained using CIFAR10 dataset, the grid search resulted in starting learning rate of 0.001 and 0.0001 for the first and second convolutional layers respectively. All these learning rates are also decayed by a rate of $0.9$.

\subsection{Tourbillon}
\label{tourbillon_section}
\begin{figure}[t!]
\begin{center}
\includegraphics[width=1.0\columnwidth]{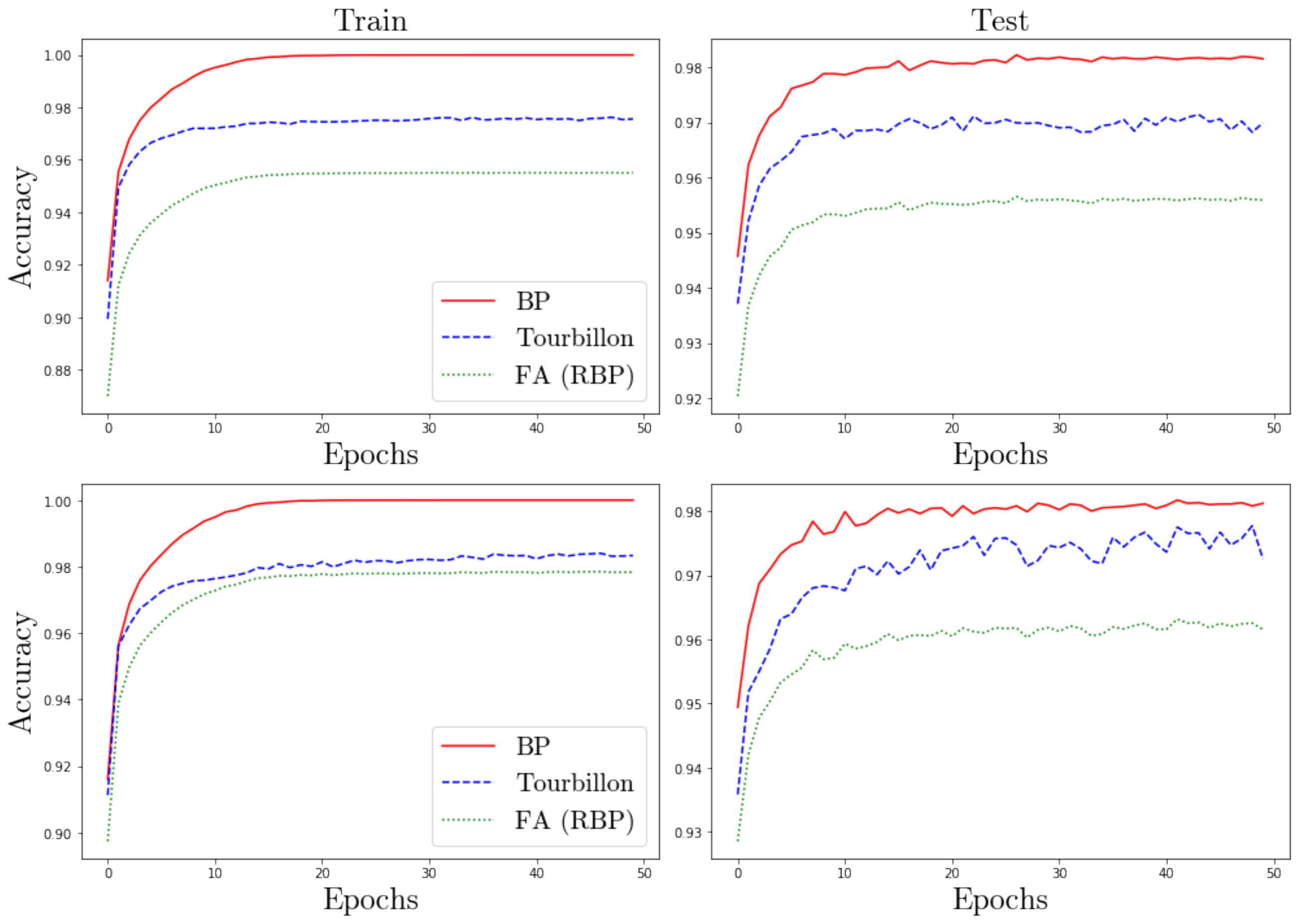}
\end{center}
\caption{Comparison of the training and test accuracy of three equivalent, fully-connected, models trained using the MNIST and Fashion MNIST datasets. Tourbillon corresponds to the Tourbillon approach. FA corresponds to the same architecture trained with feedback alignment, and BP corresponds to same architecture trained with backpropagation.}
\label{fig:mnist_fashion_tourbillon}
\end{figure}

Here we stack trained circular autoencoders to build a Tourbillon architecture. The input data to each of the circular autoencoders correspond to the hidden representations produced in the hinge layer of the previous circular autoencoder in the stack. 
\iffalse
, these Tourbillon building blocks must be trained from the bottom to the top. By this mean they would be able to generate training data for the successor building blocks. This is explained in Equation \ref{eqn:tourb} where $E$ and $D$ are representing the encoder and decoder characterized by a deep neural network trained with recirculation. $x$ is the original training data (e.g. MNIST) and $H$ shows the hidden representation of the circular autoencoder.
\be
    H_1 = E_1(x_1), \hat{x}_1 = D_1(H_1),\quad
    H_2 = E_2(H_1), \hat{H}_1 = D_2(H_2)
\label{eqn:tourb}
\ee
\fi
We perform three classification experiments using MNIST, Fashion MNIST, and CIFAR10 and compare the performance of Tourbillon with two neural networks with an equivalent architecture trained with backpropagation and random backpropagation. As shown in Figures \ref{fig:mnist_fashion_tourbillon} and the second row of Figure \ref{fig:cifar}, Tourbillon outperforms random backpropagation and is comparable to standard backpropagation, especially when using feed forward fully-connected architectures.
For MNIST and Fashion MNIST, we use the same architecture where the Tourbillon consists of two trainable circular autoencoders, the first tasked with compressing the data size from 784 to 256, and the second tasked with compressing the hinge hidden representation from 256 to 64. This is followed by a fully-connected layer with a softmax activation function to perform the final classification. Each of the circular autoencoders is trained by recirculation with the same hyperparameters used in Section \ref{circular_ae}. The top layer is initialized with Glorot initialization \citep{glorot_2010} and, after a hyperparameter grid search, its weights are optimized with a learning rate of 0.01 and a momentum of 0.8.
Similarly, for the CIFAR10 dataset, the Tourbillon architecture comprises two trainable circular autoencoders. The two circular autoencoders are similar to the circular autoencoder in Section \ref{circular_ae} for the CIFAR10 dataset.
The hinge hidden representation of the first
circular autoencoder is of size $(32 \times 32 \times 6)$. The hinge hidden representation of the second circular autoencoder is of size
$(16, 16, 6)$. The encoder component of
 these two circular autoencoders is provided by a 2D convolutional layer with \textit{tanh} activation functions and the decoder component consists of another 2D convolutional layer with a \textit{logistic} activation function. We use the same hyperparameters as we used in Section \ref{circular_ae} for training a circular autoencoder using the CIFAR10 dataset. 

\subsection{Tourbillonization}

As described in Section \ref{toubillonization}, following the Algorithm \ref{alg:tourb}, we build the physically plausible twin architecture for a U-Net autoencoder architecture that addresses all the problems mentioned in the Introduction.
In addition, we also ``tourbillonize'' a feed forward fully-connected network architecture. Results obtained on MNIST and Fashion MNIST are very similar, therefore here we report the results obtained on MNIST and CIFAR10. The U-Net architecture for the MNIST dataset comprises a two-layer encoder and a two-layer decoder. The encoder layers compress the data from 784 to 128, and then from 128 to 64. The decoder layers expand the data from 64 to 128, and then from 128 to 784. The U-Net architecture for the CIFAR10 dataset comprises two 2D convolutional layers with kernels of size $(5 \times 5)$. For both the MNIST and CIFAR10 U-Nets we use the same mini-batch size, learning rate, decay rate, momentum, and initializations as described in Section \ref{circular_ae}. After building the twin Tourbillon architectures and training their circular autoencoders by recirculation, 
we apply fine-tuning as described in Section \ref{finetune} to perform random backpropagation using the channel provided by the decoders' connections. Empirically, we found that it is necessary to have different learning rate schedules for different layers during the fine-tuning phase to ensure a form of asynchronous learning among layers(see also  
\citet{refinetti2020dynamics}).
A summary of the results comparing the various approaches is given in Table \ref{tab:finetune} (first row), and shows that the performance of the Tourbillon twin approach is comparable to backpropagation.

\iffalse
The fine-tuning
We perform random backpropagation for another 25 and 50 epochs for MNIST and CIFAR10 U-Net architectures respectively. During the fine-tuning stage, we observed that asynchronous fine-tuning of layers would improve the smoothness of the loss trajectory. We enforce this asynchronous fine-tuning by scheduling different learning rates for layers. At the beginning of the fine-tuning phase, the learning rate at the lowest layer initialized at 0.001 and decayed by a factor of $10^{-1}$ after each epoch while the learning rate of the other layers is initialized at zero. Once the learning rate of the layer $h_i$ hits the value $10^{-5}$, the learning rate of the layer $h_{i+1}$ sets to 0.001 and decays similar to the learning rate of the layer $h$. Although this is a non-adaptive way of enforcing asynchronous learning through layers, we thought that this fact is highly correlated with the dynamics of the feedback alignment described in \citet{refinetti2020dynamics}.
\fi

For the feed forward architecture, we use a three-layer network with 256, 64, and 10 hidden units. For this architecture and its Tourbillon twin, we use the same hyperparameters as in Section \ref{circular_ae}. During the fine-tuning of the Tourbillon, the learning rate schedules of each layer are similar to the rates used for the U-Net Tourbillon.
A summary of the results comparing the various approaches is given in Table \ref{tab:finetune} (second row), and shows again that the performance of the Tourbillon twin approach is comparable to backpropagation.

\begin{table*}[h]
%\normalsize
\small
\centering
\caption{Performance comparison of networks trained with backpropagation vs their physically plausible Tourbillon twin. For the U-Net experiment, the numbers represent the mean-squared reconstruction error. For the feed forward fully-connected architecture, the numbers represent the classification error on the test dataset. For every reported number, the standard deviation is also shown in the parenthesis. For the sake of brevity, FC stands for fully-connected network.}
 \vspace{3mm}
 \begin{tabular}{cccccc} 
 \toprule
  Architecture & &\multicolumn{2}{c}{MNIST} & \multicolumn{2}{c}{CIFAR10}\\
  \cmidrule(lr){3-4}
  \cmidrule(lr){5-6}
  & &Train & Test & Train & Test\\
 \midrule
 \multirow{2}{*}{U-Net} &BP&0.0031 (5.8 e-05)& 0.0031 (5.8 e-05)&0.0024 (5.8 e-05)&0.0026 (e-04)\\
 &Twin&0.0111 (e-04)&0.0113 (2e-04)&0.0733 (3.2e-04)&0.0783 (3.2e-04)\\
 \cmidrule(lr){1-6}
  \multirow{2}{*}{FC} &BP&1.04 (0.20)&1.18 (0.20)&-&-\\
  &Twin&5.21 (0.40)&5.82 (0.40)&-&-\\
 \bottomrule
 \end{tabular}
 \label{tab:finetune}
\end{table*}

\section{Discussion and Conclusion}
In a physical neural system, implementation of backpropagation is faced with a number of
significant challenges described in the Introduction. Here we have presented a systematic approach we call Tourbillon for deriving physically plausible architectures addressing most if not all of them. At its core, Tourbillon relies on a stack of circular autoencoders trained by recirculation. Tourbillon architectures are modular and can be trained in large part in self-supervised mode using learning rules that are local in both space and time, without the need for weight transport. 

The name Tourbillon captures the turbulent topology of the architecture. Moreover, in horology, a tourbillon is an addition to the mechanics of a watch escapement to increase its accuracy. While we do not claim to have increased accuracy, we have shown that the Tourbillon approach achieves results that are comparable to backpropagation, at least on MNIST, Fashion MNIST, and CIFAR10.

In the development of Tourbillon, several
issues have been identified that will require additional research. The first one is to study whether local learning algorithms can be developed to improve the fine-tuning phase of Tourbillon architectures where the encoder components of the circular autoencoders have multiple hidden layers. Something along the lines of recirculating the RBP error may be a possibility. 

The second one is to develop local learning algorithms for the recursive Tourbillon approach, starting with Tourbillions of Tourbillons. In particular, one would like to know whether the fine-tuning phase is necessary and how to carry it, perhaps using a fast form of recirculation in the smaller circular autoencoders and a slower form of recirculation in the larger circular autoencoder. 

The third one is the issue of convolutions that were only incrementally addressed here by showing that Tourbillon with convolutions does better than RBP, but lags behind BP.
When convolutions are used, neurons in a convolution layer must have identical incoming weights, which may not be easy to realize in a physical neural system. The most plausible approach to address the convolution problem in a physical system is to first initialize the weights in similar, but not identical, fashion across the entire convolutional layer, using for instance normal weights with small standard deviation. This ensures that all the weights are similar, without enforcing them to be identical. Second, during training, each convolution neuron in the layer learns independently of the other convolution neurons, without enforcing any kind of rigid weight sharing. What is essential, however, is to ensure that all the convolution neurons in the layer see exactly, or approximately, the same training data in aggregate. This is easily achieved through data augmentation by translating each training image in all possible directions, something that may happen automatically in the real world due to moving objects, or head/eye motions. With this data augmentation, the weights of the convolution neurons remain similar throughout training, since they are trained on the same data, without any exact weight sharing \citep{ott2020learning}. This approach ought to be tried in Tourbillon. 
\iffalse
    \item An alternative possibility which, however, is not very physically plausible is to train only one neuron in the convolution layer using the same augmented data as above and, at the end of the training, copy the weights of this neuron to all the other neurons in the same layer. 
    The data augmentation is still necessary to ensure that the training neurons see all the kinds of possible data while avoiding the averaging effect described above. While not physical, this approach is easy to implement in current software libraries and yields exact weight sharing. 
\end{enumerate}
\fi

Tourbillon provides a framework for investigating these and other issues and broadens the horizon of ongoing research into 
physically plausible deep learning.

\iffalse
While these directions are being investigated, TOur proposed architecture, results, and implementations are expected to establish baselines and broaden the horizon of ongoing research in the field of physically plausible deep learning.
\fi

\bibliographystyle{unsrtnat}
\bibliography{main}

\section{Appendix}
\label{appendix}
In this section, we first provide more explanation on the new learning rule introduced in Equation \ref{eqn:new_lr}. Then we include evaluation plots corresponding to: (1) The performance of the circular autoencoder and Tourbillon models trained using the CIFAR10 dataset and their comparison to models trained using random backpropagation and backpropagation; (2) The representation power of Tourbillon models by visualizing the reconstructed images; and (3) The process of tourbillonization.

\begin{figure}[h!]
\begin{center}
\includegraphics[width=1.0\columnwidth]{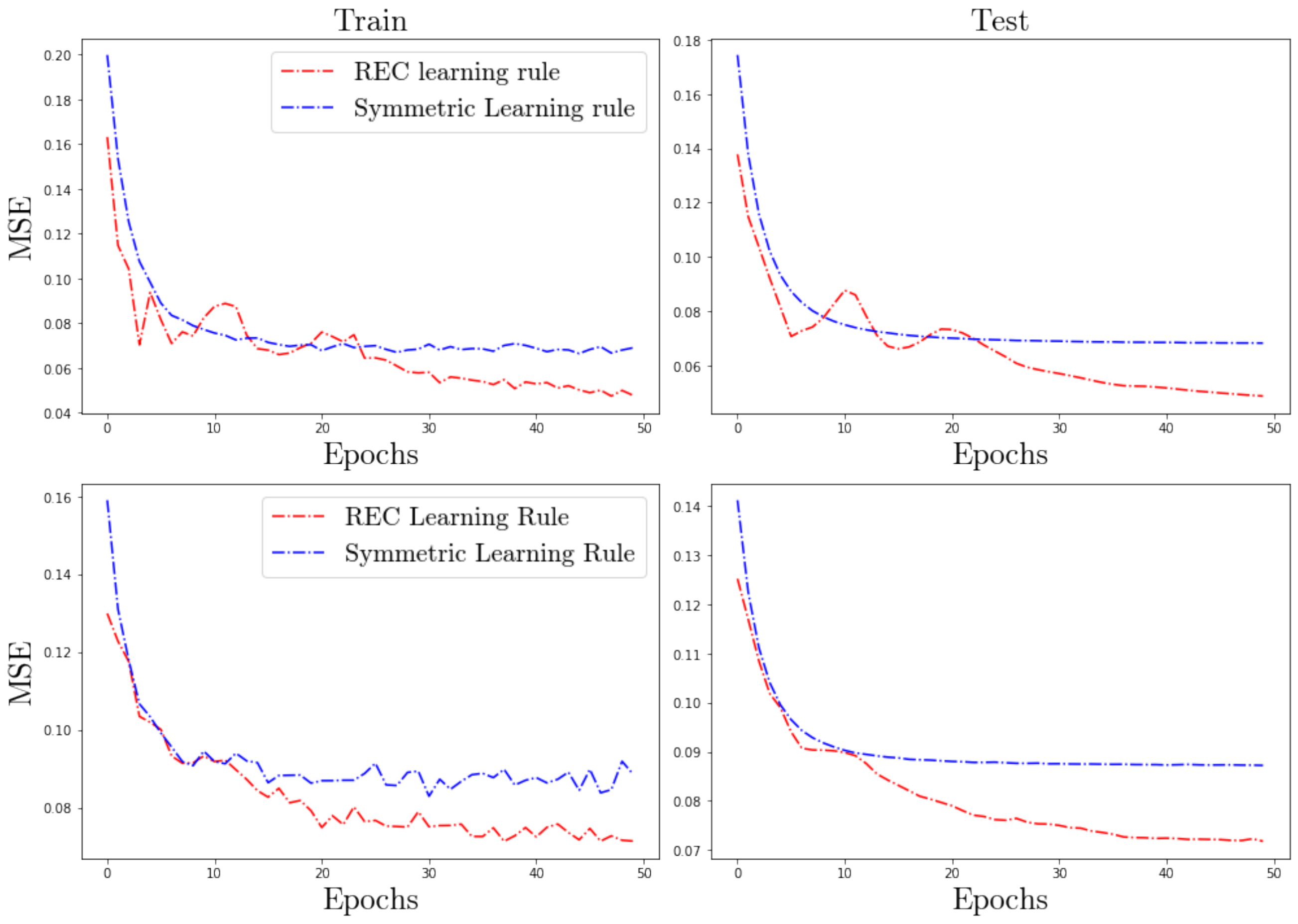}
\end{center}
\caption{The different behavior of the two learning rules introduced in Section \ref{learninig}. The symmetric learning rule corresponds to Equation \ref{eqn:new_lr} and the REC learning rule corresponds to Equation \ref{eqn:rec}. While both demonstrate similar performance in training models using MNIST (first row) and Fashion MNIST (second row), the symmetric learning rule has smoother trajectories.}
\label{fig:learning_rules_traj}
\end{figure}

\begin{figure}[h!]
\begin{center}
\includegraphics[width=1.0\columnwidth]{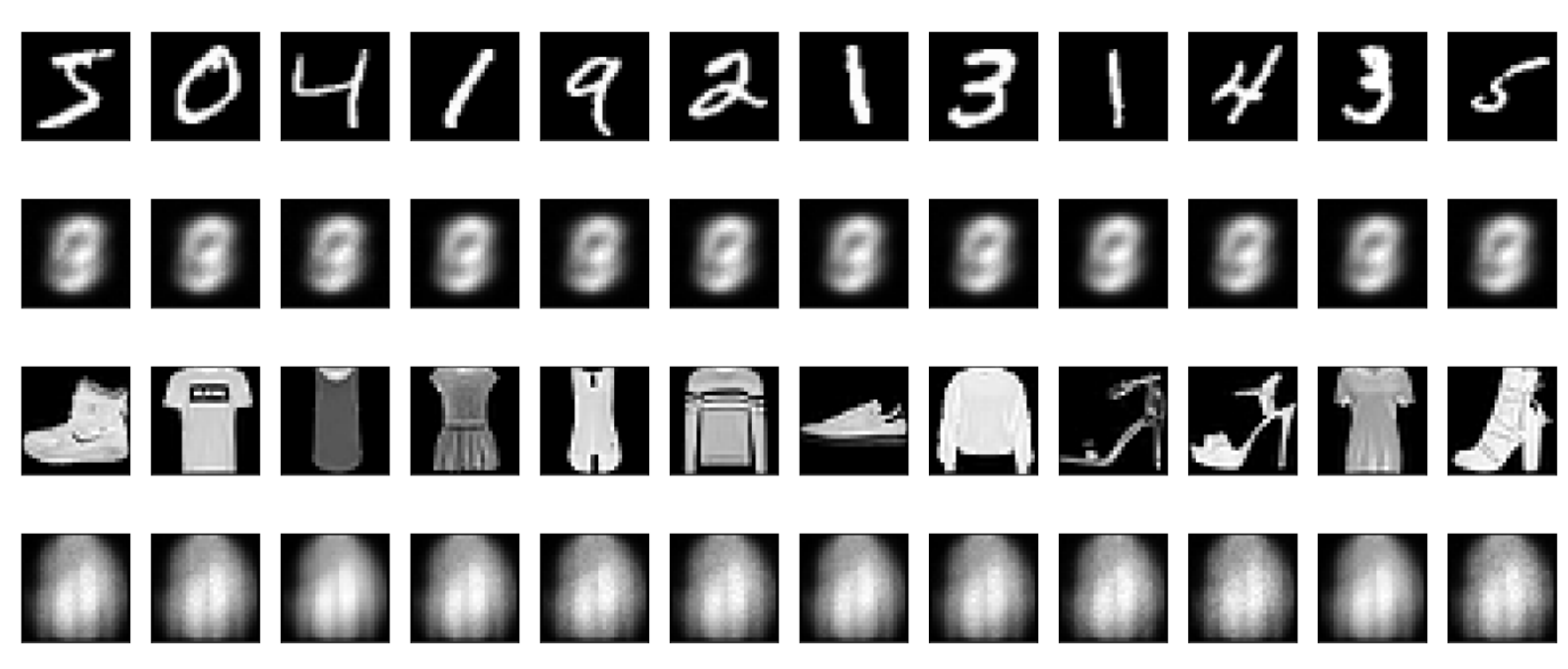}
\end{center}
\caption{Examples of reconstructed images from the MNIST and Fashion MINST datasets using the circular autoencoder trained with the symmetric learning rule. There is a mode-collapse effect and the model reconstructs the mean of the data.}
\label{fig:mnist_reconst_new}
\end{figure}

Here we rewrite the symmetric learning rule we introduced earlier:
\be
\Delta w_{ij} =-\eta
[O_i(0)-O_i(1)]^{post}[O_j(0)-O_j(1)]^{pre}
\label{eqn:new_lr2}
\ee
This learning rule has symmetric terms for the pre- and post-synaptic activation differences.
Although the loss trajectory is smoother than when using the main recirculation rule (Figure \ref{fig:learning_rules_traj}), the algorithm becomes trapped into a mode-collapse state where the reconstructed images are the mean of the entire dataset (Figure \ref{fig:mnist_reconst_new}). Further studies of this learning rule and its mode collapse are left for future work.

\iffalse
Due to the cyclic structure of circular autoencoders, the data circulate through the model in multiple time steps where neurons produce different activations. Assuming the time frame between two consecutive cycles is short, the difference between two consecutive activations $[O_j(1)-O_j(0)]$ can be interpreted to be proportional to the rate of the neuron activation in time. In other words, if $O_j(1) \geq O_j(0)$ the neurons are activating in that time step. Traditionally, the neurons' activation rate is assumed to be proportional to the sigmoidal transformation of the neurons' output, however, the term $[O_j(0)-O_j(1)]$ can be another measure of the activation rate of neurons in a neural network.
\fi

\iffalse
We consider the weight update as the multiplication of this term for pre-synaptic and post-synaptic neurons. In other words, this definition implies that the connection between two neurons must be strengthened if neurons' activation are correlated and must be weakened otherwise (i.e. neurons that fire together, wire together). However, despite the interesting interpretation, we found problems in training circular autoencoders using this symmetric learning rule.
\fi

\begin{figure}[h!]
\begin{center}
\includegraphics[width=1.0\columnwidth]{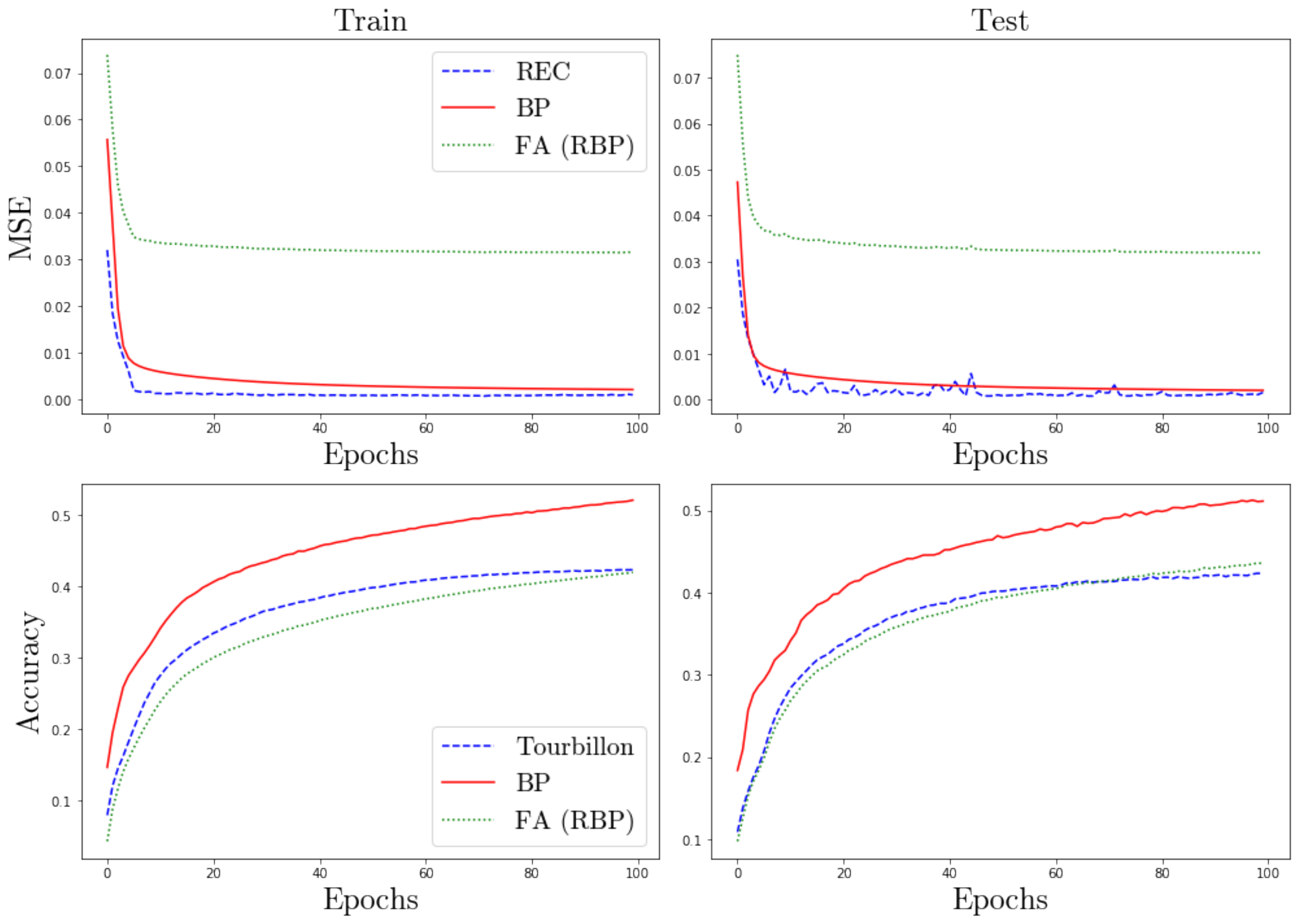}
\end{center}
\caption{First row: Comparison of the mean-squared reconstruction loss between circular autoencoder with recirculation, feedback alignment, and backpropagation. Second row: Comparison of the training and test accuracy for Tourbillon, feedback alignment, and backpropagation using the CIFAR10 dataset.}
\label{fig:cifar}
\end{figure}

The first row of Figure \ref{fig:cifar} represents the training and test performance of a circular autoencoder trained by recirculation using the CIFAR10 dataset and its comparison to an autoencoder with similar architecture trained with random backpropagation and backpropagation.

\begin{figure}[h!]
\begin{center}
\includegraphics[width=\columnwidth]{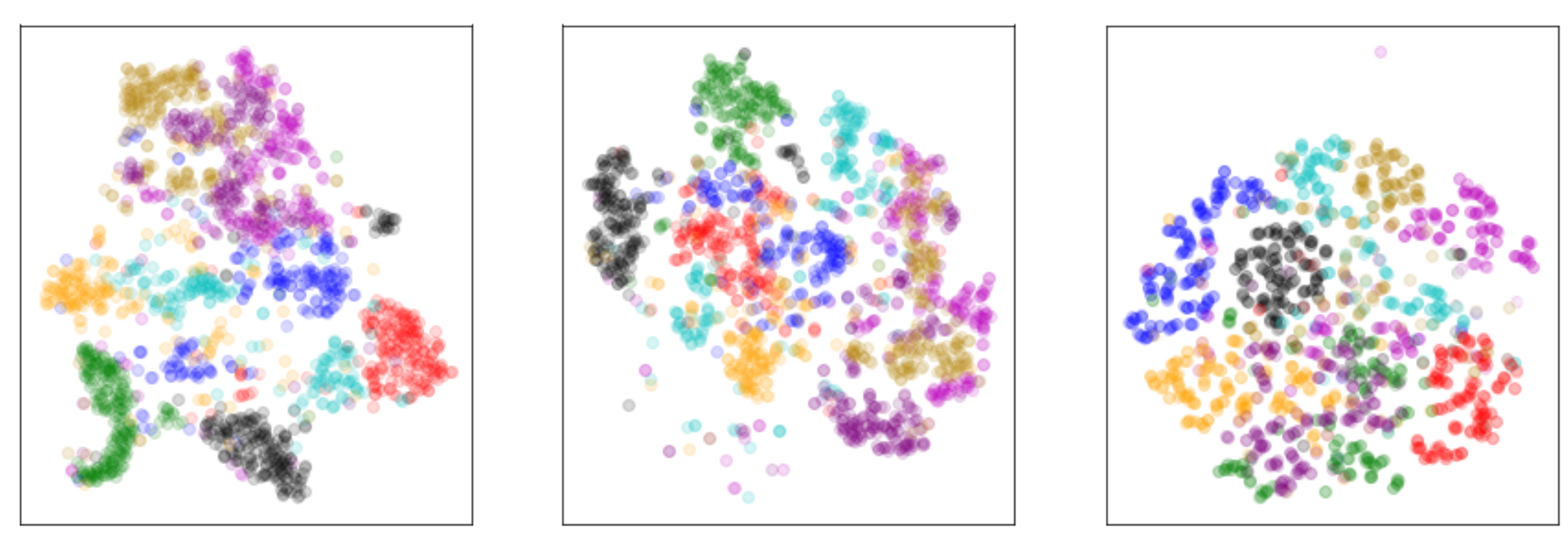}
\end{center}
\caption{From left to right, the plots correspond to the 2D visualization of 2000 random samples from the MNIST, Fashion MNIST, and CIFAR10 datasets.}
\label{fig:tsne}
\end{figure}

\begin{figure}[h!]
\begin{center}
\includegraphics[width=1.0\columnwidth]{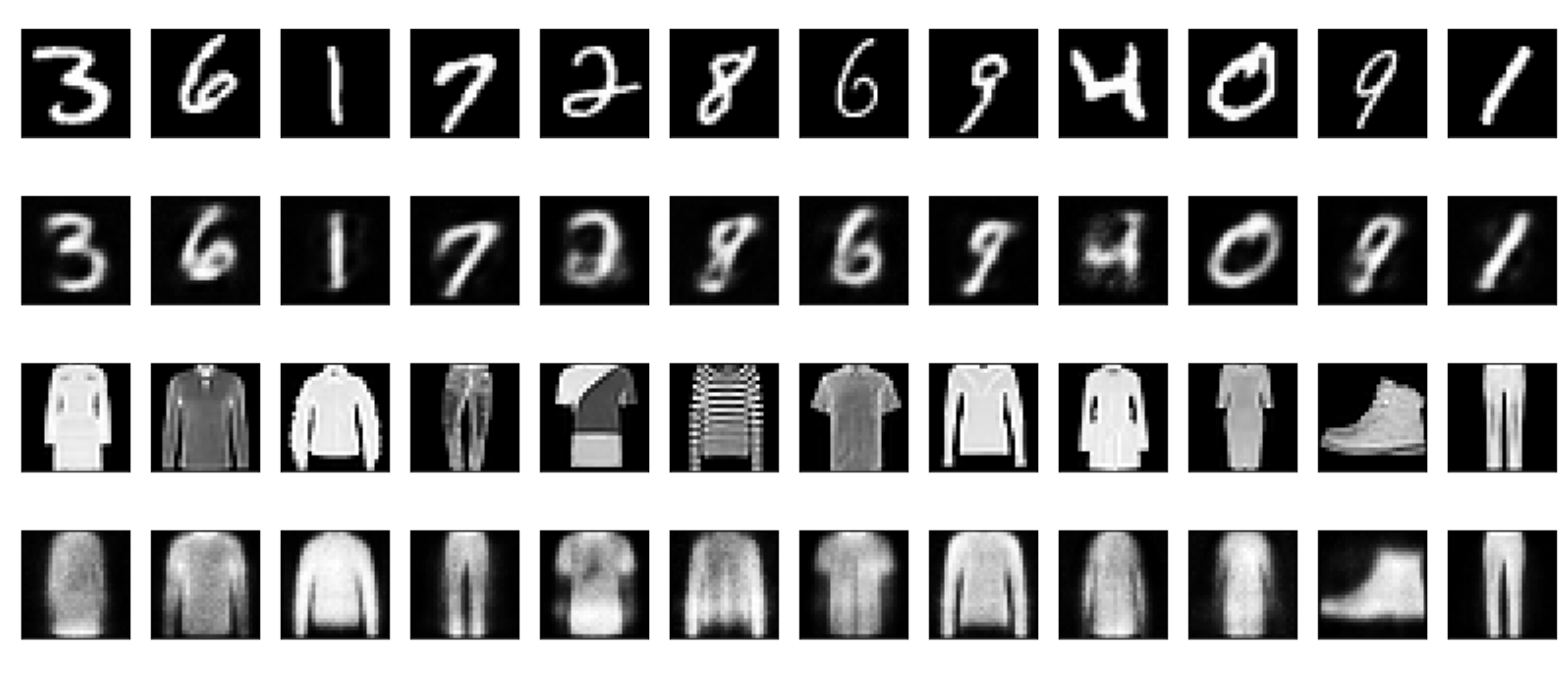}
\end{center}
\caption{Examples of reconstructed images from the MNIST and the Fashion MNIST datasets using Tourbillon.}
\label{fig:tourb_reconst}
\end{figure}

Figure \ref{fig:tourb_reconst} shows the reconstructed images using a stack of two circular autoencoders. We also use the tSNE \citep{van2008visualizing} dimensionality reduction method in Figure \ref{fig:tsne} to visualize the output of circular autoencoders used in Section \ref{tourbillon_section}. These two figures show the representational power of the Tourbillon building blocks.

\begin{figure}[h!]
\begin{center}
\includegraphics[width=1.0\columnwidth]{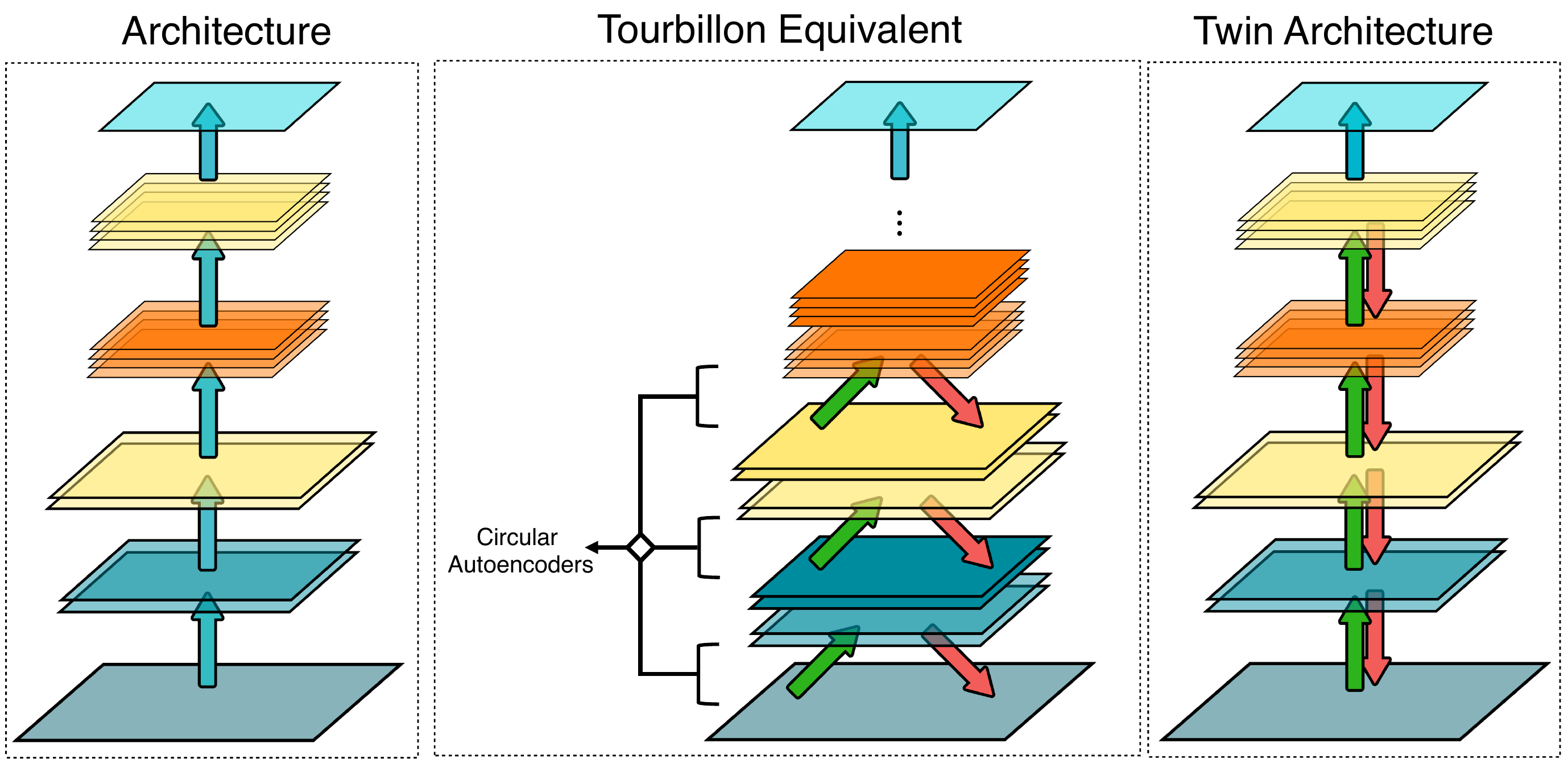}
\end{center}
\caption{Schematic visualization of the ``tourbillonization'' process. Left: A given neural network model where the blue arrows represent the weights that are trained by backpropagation. Middle: The Tourbillon equivalent of the given neural network. Each of the intermediate layers of the given network are replaced with a circular autoencoder that has a one-layer encoder and a one-layer decoder. Both red and green arrows show the weights that are trained by recirculation. Right: The final tourbillonized model where the forward weight (green arrows) can be fine-tuned using the decoder's channel (red arrows) by random backpropagation.}
\label{fig:recirculation_mnist}
\end{figure}

\end{document}